\documentclass[journal]{IEEEtran}

\usepackage{cite}
\usepackage{amsmath,amssymb,amsfonts}
\usepackage{algorithmic}
\usepackage{graphicx}
\usepackage{textcomp}
\usepackage{booktabs}
\usepackage[hidelinks]{hyperref}

\usepackage{array,multirow}
\usepackage{float}

\begin{document}

\newcommand{\orcid}[1]{\href{https://orcid.org/#1}{\includegraphics[height = 2ex]{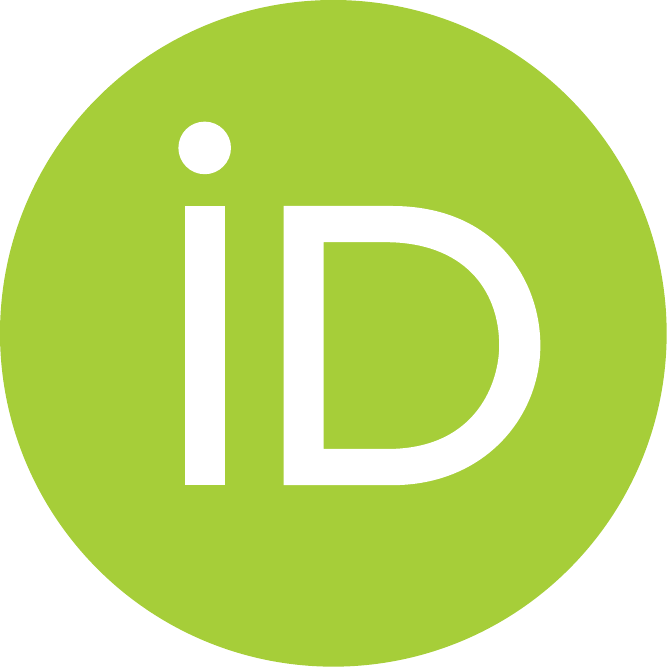}}}

\title{Semantic Segmentation for Autonomous Driving: Model Evaluation, Dataset Generation, Perspective Comparison, and Real-Time Capability}

\author{
    \IEEEauthorblockN{\uppercase{Senay Cakir} \orcid{0000-0003-3970-3072}, \uppercase{Marcel Gau\ss} \orcid{0000-0003-2472-5981}, \uppercase{Kai H\"appeler} \orcid{0000-0002-9654-8657}, \uppercase{Yassine Ounajjar}  \orcid{0000-0002-0701-0944}, \\\uppercase{Fabian Heinle} \orcid{0000-0002-3933-3377}, \uppercase{and Reiner Marchthaler} \orcid{0000-0001-9108-3072}}
    \IEEEauthorblockA{\\\IEEEauthorrefmark{1} Esslingen University of Applied Sciences, Flandernstra{\ss}e 101, 73732 Esslingen am Neckar, GERMANY}
}

\markboth{arXiv.org, open-access archive, computer science, July~2022}%
{}

%+++++++++++++++++++++++++++++++++++++++++++++++++++++++++++++++++++++++++++++++++++++++++++
\maketitle 																														% make the title area
%+++++++++++++++++++++++++++++++++++++++++++++++++++++++++++++++++++++++++++++++++++++++++++

% As a general rule, do not put math, special symbols or citations in the abstract or keywords.
\begin{abstract}
Environmental perception is an important aspect within the field of autonomous vehicles that provides crucial information about the driving domain, including but not limited to identifying clear driving areas and surrounding obstacles. Semantic segmentation is a widely used perception method for self-driving cars that associates each pixel of an image with a predefined class. In this context, several segmentation models are evaluated regarding accuracy and efficiency. Experimental results on the generated dataset confirm that the segmentation model FasterSeg is fast enough to be used in realtime on low-power computational (embedded) devices in self-driving cars. A simple method is also introduced to generate synthetic training data for the model.
Moreover, the accuracy of the first-person perspective and the bird's eye view perspective are compared. 
For a $320 \times 256$ input in the first-person perspective, FasterSeg achieves $65.44\,\%$ mean Intersection over Union (mIoU), and for a $320 \times 256$ input from the bird's eye view perspective, FasterSeg achieves $64.08\,\%$ mIoU. Both perspectives achieve a frame rate of $247.11$ Frames per Second (FPS) on the NVIDIA Jetson AGX Xavier. Lastly, the frame rate and the accuracy with respect to the arithmetic 16-bit Floating Point (FP16) and 32-bit Floating Point (FP32) of both perspectives are measured and compared on the target hardware.
\end{abstract}


% For peer review papers, you can put extra information on the cover page as needed:
% \fi
%
% For peerreview papers, this IEEEtran command inserts a page break and creates the second title. It will be ignored for other modes.
\IEEEpeerreviewmaketitle

%+++++++++++++++++++++++++++++++++++++++++++++++++++++++++++++++++++++++++++++++++++++++++++
%Intro section
\section{Introduction}

\label{sec:introduction}
\IEEEPARstart{A}{utonomous} vehicles have a variety of different sensor systems onboard to detect obstacles, lanes, free parking spaces, etc. \cite{k11}, \cite{k12}. A frequently applied technique in this field is image segmentation, which uses camera images to classify each pixel. The predicted images can be used to plan the vehicle's behavior and avoid collisions\cite{k13}, \cite{k14}. This work was conducted in the context of the Carolo-Cup, a student competition providing student teams with a platform for the design and implementation of autonomous Radio Controlled (RC) vehicles. They must accomplish various driving tasks such as parking or overtaking in an imitated environment containing obstacles, intersections, parking spaces, and more.
Furthermore, RC vehicles use embedded hardware to run the sensing, planning, control algorithms, etc. The algorithms must therefore run in realtime so that the vehicle can drive smoothly and reliably. \cite{k15} \\
To classify the pixels of the images delivered by the camera built on top of the vehicle, several image segmentation models were evaluated. A dataset representing the imitated environment is required to train the segmentation neural network. In this context, synthetic images generated with a simulation are combined with real images of the Carolo-Cup environment to compose the training dataset. Supervised learning is used in this work because each image of the dataset has its corresponding ground truth.
The motivation of this work is to generate a dataset that mainly contains synthetic data to avoid high labeling effort. Thus, the routes can be generated in a simulation and must not be replicated. Moreover, a state-of-the-art image segmentation model is applied in realtime on a comparatively slow embedded hardware. 
Additionally, the potential of the bird's eye view perspective is examined. Both the overall accuracy mean Intersection over Union (mIoU) and the accuracy Intersection over Union (IoU) of each class are then investigated more closely. This paper attempts to answer four main questions:

\begin{itemize}
    \item Which image segmentation model is fast and accurate enough for the Carolo-Cup?
    \item How to easily generate labeled synthetic data?
    \item Is the bird's eye view perspective a better alternative compared to the first-person perspective?
    \item What impact does the 16-bit Floating Point (FP16) and the 32-bit Floating Point (FP32) arithmetic have on the model accuracy and the real-time capability? 
\end{itemize} 
This paper is organized as follows. First, different segmentation models are evaluated to find a suitable option for this work. Secondly, a method for generating labeled synthetic data is described. Lastly, two different experiments are conducted, using the selected segmentation model trained with the generated data. The first experiment examines the accuracy of two models trained with data from two different perspectives: the first-person and the bird's eye view perspective. The second experiment explores the real-time capability and accuracy regarding the arithmetic FP16 and FP32 of both models. The intended contributions of this study are the following:
\begin{itemize}
    \item The development of a simple, yet effective method to generate synthetic data representing an imitated environment for autonomous vehicles.
    \item Exploring the possibility of executing semantic segmentation on low-power embedded devices using images from the bird's eye view perspective.
\end{itemize}

%+++++++++++++++++++++++++++++++++++++++++++++++++++++++++++++++++++++++++++++++++++++++++++
%++++++++++++++++++++++++++++++++++++++++++++++++++++++++++++++++++++++++++++++++++related work section

\section{Related work}
The following section describes related work which is relevant. First, the fundamentals of synthetic data generation are introduced. In the second paragraph, an overview of real data sources is provided. Then, various image segmentation models are evaluated and compared in terms of accuracy and frame rate. Finally, the chosen image segmentation model is further described.
%---------------------------------------------------------------------
\subsection{ROAD GENERATION AND SIMULATION}
\label{subsec:Roadgen and Simu}
Gazebo is chosen as the simulation environment to replicate realistic driving scenes. The synthetic routes used in Gazebo can be generated as images using a road generator provided by \cite{k2a1}. These images can be directly rendered in the simulation environment. To create various routes, the road generator is extended with the objects listed in Fig.~\ref{fig:annotation definitions}. Furthermore, the generator is customized to create equivalent annotated routes \cite{k2a2}.
%---------------------------------------------------------------------
\subsection{SOURCES OF REAL DATA}
\label{subsec:Sources of real data}
In addition to the generated synthetic data, images from real routes in imitated environments are used. These real images are provided by various research teams such as Spatzenhirn (University of Ulm) \cite{k2b1}, ISF L\"owen (Technical University of Braunschweig) \cite{k2b2}, KITcar (Karlsruhe Institute of Technology) \cite{k2b3}, and it:movES (Esslingen University of Applied Sciences) \cite{k2b4}. Different environments offer a relatively high diversity of real images which can be very useful for training and testing an image segmentation model. The images are recorded using different cameras.
%---------------------------------------------------------------------
\subsection{EVALUATION OF STATE-OF-THE-ART IMAGE SEGMENTATION MODELS}
Image segmentation is an important part of visual perception systems for autonomous vehicles. It can be described as separating an image into any segments. Image segmentation can be divided into semantic segmentation and instance segmentation. Semantic segmentation refers to the process of assigning a label to each pixel of a picture. Instance segmentation extends the semantic segmentation scope further by detecting each instance of the object within the image and delineating it with a bounding box or segmentation mask. \cite{k30} \\
To interpret the images of the vehicle's environment, different instance and semantic segmentation models were evaluated. The goal of the evaluation is to find a model that achieves a high inference frame rate measured in Frames per Second (FPS) while concurrently maintaining high detection accuracy. \\
Table \ref{tab:evalInstanceSeg} lists the frame rate as well as the accuracy of some state-of-the-art instance segmentation models on different GPUs. The models were rated using the MS COCO benchmark dataset \cite{k2c1}, and the Average Precision (AP) was used as an accuracy metric. Table \ref{tab:evalSemanticSeg} lists some state-of-the-art semantic segmentation models rated with the Cityscapes benchmark dataset \cite{k2c2}. The $mIoU$ is used to measure the models' accuracies. The $mIoU$ is defined as follows:
\begin{equation}
    mIoU = \frac{1}{N} \sum_{n=1}^{N} IoU = \frac{1}{N} \sum_{n=1}^{N} \frac{TP}{(TP+FP+FN)}
\end{equation}
where $TP$, $FP$, and $FN$ are the numbers of true positive, false positive, and false negative pixels determined over the whole test set. $N$ is the number of the defined classes. \cite{k31} \\
The evaluation demonstrates that semantic segmentation models achieve higher frame rates and accuracy values using higher image resolution. Therefore semantic segmentation is chosen due to it's likelihood to run successfully in realtime on the used hardware. As Table \ref{tab:evalSemanticSeg} indicates, models such as BiSeNetV2 \cite{k28} or FasterSeg \cite{k29} achieve a comparatively high frame rate, and the accuracy is only slightly smaller compared to other models like U-HarDnet-70 \cite{k25} or SwiftNetRN-18 \cite{k27}. To reach real-time capability and good accuracy on the used embedded hardware, FasterSeg is chosen as a segmentation model for this work. FasterSeg attains the highest frame rate of the compared models with 163.9 FPS and a relatively good $mIoU$ with $71.5\,\%$.
%---------------------------------------------------------------------
\begin{table}
\caption{Evaluation of the instance segmentation models.}
\setlength{\tabcolsep}{3pt}
\label{tab:evalInstanceSeg}
    \begin{tabular}{p{66pt}p{50pt}p{25pt}p{25pt}p{47pt}}\toprule    
        \textbf{Model}	       &\textbf{Resolution} &\textbf{AP [$\%$]} &\textbf{FPS} &\textbf{GPU} \\\midrule
        BlendMask-512          & 550 x 550          & 36.8              & 25.0        & GeForce GTX \\
        \cite{k21}             &                    &                   &             & 1080 Ti     \\
        YOLACT-550++ 	       & 550 x 550	        & 34.1	            & 33.5        & GeForce GTX \\
        \cite{k22}             &                    &                   &             & Titan Xp    \\
        CenterMask		       & 580 x 600	        & 38.3              & 35.0        & GeForce GTX \\
        \cite{k23}             &                    &                   &             & Titan XP\\
        YOLACT-550 \cite{k24}  & 550 x 550	        & 29.8	            & 33.5        & GeForce GTX \\
                               &                    &                   &             & Titan Xp    \\\bottomrule
    \end{tabular}
\end{table}

\begin{table}
    \caption{Evaluation of the semantic segmentation models.}
    \setlength{\tabcolsep}{3pt}
    \label{tab:evalSemanticSeg}
        \begin{tabular}{p{66pt}p{50pt}p{25pt}p{25pt}p{47pt}} \toprule 
            \textbf{Model}           &\textbf{Resolution} &\textbf{mIoU}   &\textbf{FPS} &\textbf{GPU} \\
                                     &                    &\textbf{[$\%$]} &             &             \\\midrule
            U-HarDnet-70 \cite{k25}  & 1024 x 2048        & 75.9           & 53.0        & GeForce GTX \\
            \cite{k26}               &                    &                &             & 1080 Ti     \\       
            SwiftNetRN-18 \cite{k27} & 1024 x 2048        & 75.5           & 39.9        & GeForce GTX \\
                                     &                    &                &             & 1080 Ti     \\       
            BiSeNetV2 \cite{k28}     & 1024 x 2048        & 72.6           & 156.0       & GeForce GTX \\
                                     &                    &                &             & 1080 Ti     \\   
            FasterSeg \cite{k29}     & 1024 x 2048        & 71.5           & 163.9       & GeForce GTX \\
                                     &                    &                &             & 1080 Ti     \\ 
            BiSeNet \cite{k210}      & 1024 x 2048        & 68.4           & 105.8       & Titan Xp    \\\bottomrule 
        \end{tabular}
\end{table}
%---------------------------------------------------------------------
\subsection{FASTERSEG}
\begin{figure}[t!]
    \includegraphics[width=.46\textwidth]{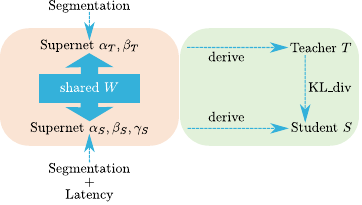}
    \caption{During search (left), the co-searching framework optimizes two architectures, and during training from scratch (right), it distills from a complicated teacher to a light student using KL \cite{k29}.}
	\label{fig:FasterSeg}
\end{figure}

FasterSeg is an automatically designed neural network for semantic segmentation that was discovered by a novel Neural Architecture Search (NAS) framework. This network has an efficient search space with multi-resolution branching. With the intention of reaching an optimal trade-off between low latency and high accuracy, this network integrates a fast inference speed, competitive accuracy and a fine grained latency regularization.
Knowledge Distillation (KL) is a technique used by FasterSeg to transfer knowledge from a large complex network (teacher $T$) to a much smaller network (student $S$). Fig. \ref{fig:FasterSeg} shows that the searched teacher is a sophisticated network based on the same search space and supernet weights $W$ as the student. FasterSeg uses two sets of architectures in one supernet to simultaneously search for a complicated teacher $(\alpha_T, \beta_T )$ and a lightweight student $(\alpha_S , \beta_S , \gamma_S)$. FasterSeg training can be broken down into four stages:
\begin{itemize}
    \item Search the architecture
    \item Pre-train the supernet
    \item Pre-train the teacher network
    \item Pre-train the student network
\end{itemize}
In all experiments conducted the supernet is pre-trained for 20 epochs without changing the architecture parameters. Then the architecture search is done for 30 epochs. The epoch values are the same as used for the search experiments run by the FasterSeg developing team. \cite{k29}

\begin{figure*}[t!]
    \includegraphics[width=\textwidth]{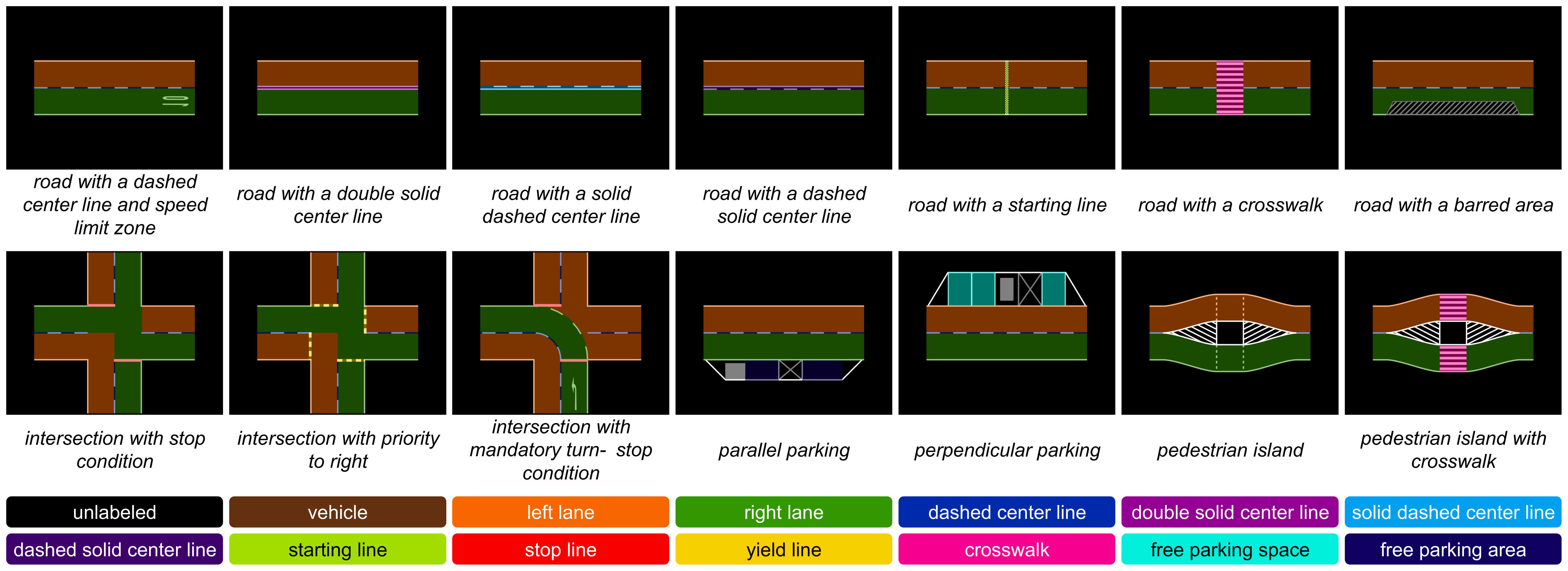}
    \caption{Definition of the existing objects. Each color represents the corresponding class, which should be recognized.}
	\label{fig:annotation definitions}
\end{figure*}

\begin{figure*}[t!]
    \includegraphics[width=\textwidth]{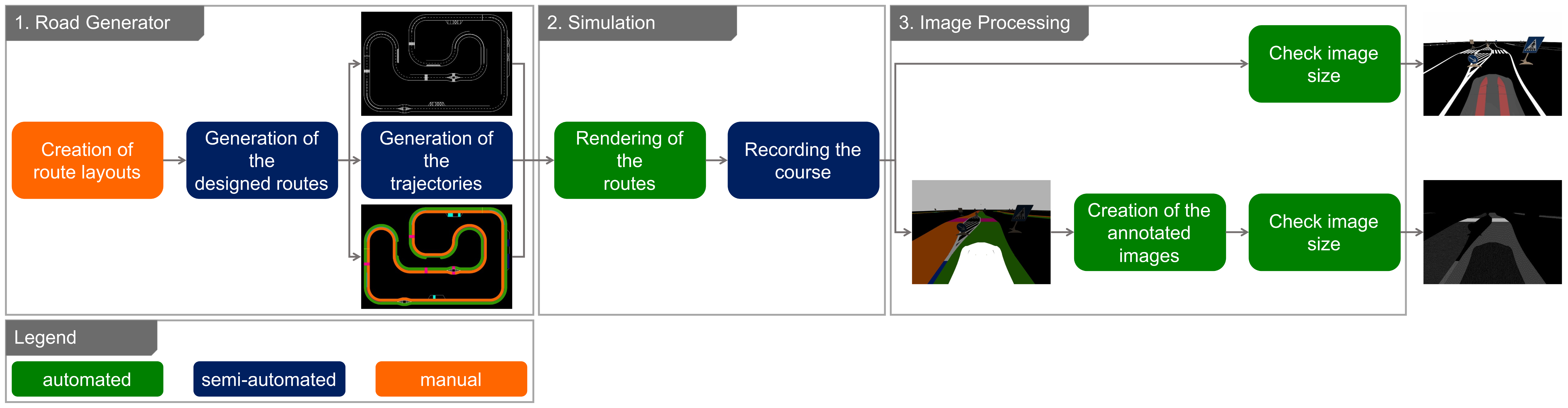}
    \caption{System overview of the synthetic data generation.}
	\label{fig:system overview}
\end{figure*}

%+++++++++++++++++++++++++++++++++++++++++++++++++++++++++++++++++++++++++++++++++++++++++++
% Proposed approach section
\section{Proposed approach}
The definitions of the labels with various scenarios of the imitated environment are shown in Fig.~\ref{fig:annotation definitions}. In this section, the generation of the labeled data is described. Additionally, a method to transform the images into a bird's eye view perspective is presented. 
%---------------------------------------------------------------------
\subsection{SYNTHETIC DATA GENERATION}
\label{subsec:SynData Generation}
The process of synthetic data generation is shown in its entirety in Fig.~\ref{fig:system overview}. This process is divided into three fields: the road generator, the simulation, and the image processing. Furthermore, the automation level of each task within the fields is visualized with a corresponding color. In the following, the tasks of each main field of the figure are described.\\

\subsubsection{Road Generator}
\hfill\\
Synthetic data generation starts with the creation of a route layout. High diversity and different constellations are essential for accurate predictions. Therefore, various configurations of parking zones, intersections, center line types, missing lines, objects, and curves with different radii and angles must be created. A raw and an annotated route are automatically generated using the designed layout. The road generator also creates x- and y- coordinates, as well as the yaw angle. This represents the spatial orientation for the trajectory of the simulated vehicle. The coordinates run along the center of the right lane. Special driving maneuvers, such as overtaking, parking, or crossing the intersection from different directions, must be added manually.\\

\subsubsection{Simulation}
\begin{figure}[t!]
    \includegraphics[width=.48\textwidth]{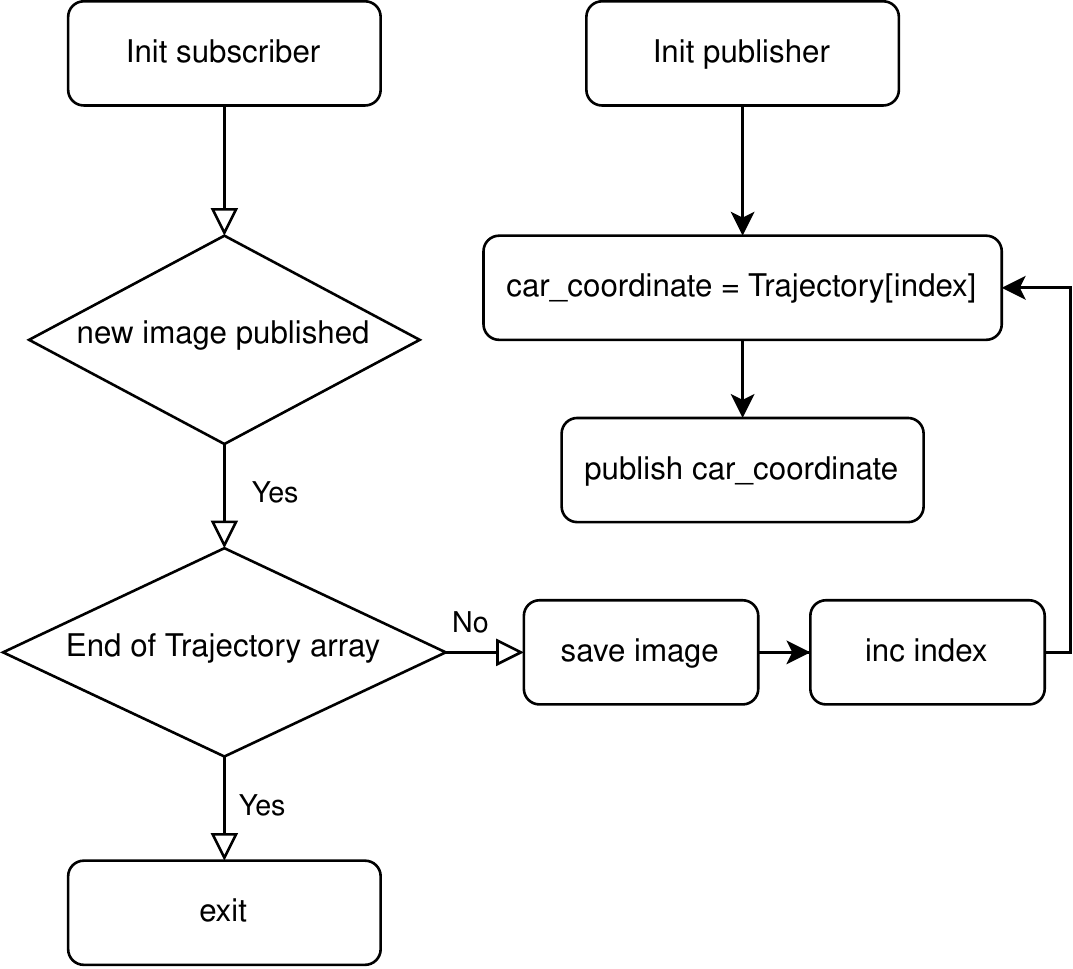}
    \caption{Flow chart of the publishing process of the trajectory in Gazebo.}
	\label{fig:flow_chart_simu}
\end{figure}

\hfill\\
As described in Section \ref{subsec:Roadgen and Simu}, Gazebo is selected as the simulation environment to generate synthetic training data for this work. The simulator produces realistic first-person perspective image sequences that replicate real driving scenarios. To achieve this, two different virtual vehicles are rendered in the simulator, each one driving on a different route created by the road generator. The first vehicle is driving on the raw route, while the second one is assigned the colored route. Due to the fact Gazebo is based on Robotic Operating System (ROS), the architecture and therefore the behavior of both vehicles are similar. The virtual vehicles were built based on the RC vehicle used at Esslingen University. After rendering both vehicles and routes in the simulation environment, the trajectory generated by the road generator is published using ROS. The publishing of the trajectory is executed for both vehicles at the same time. The camera topics are then recorded to produce two synchronized sequences of raw and colored images. These images are finally sent to the next stage for processing. Fig.~\ref{fig:flow_chart_simu} shows a flow chart of the trajectory publishing process.\\

\subsubsection{Image Processing}
\hfill\\
To process the annotations, FasterSeg requires an 8-bit grayscale image where each pixel contains the class ID. Therefore, each pixel of the colored image is replaced by the corresponding class ID using a lookup table. The result is an 8-bit grayscale image with IDs representing each class. Additionally, a Region of Interest (ROI) is set to exclude undetectable objects near the horizon. FasterSeg also requires the image height and width to be divisible by 64. The generated images are thus checked and eventually downscaled.
%---------------------------------------------------------------------
\subsection{REAL DATA GENERATION}
\begin{figure}[t!]
    \includegraphics[width=.48\textwidth]{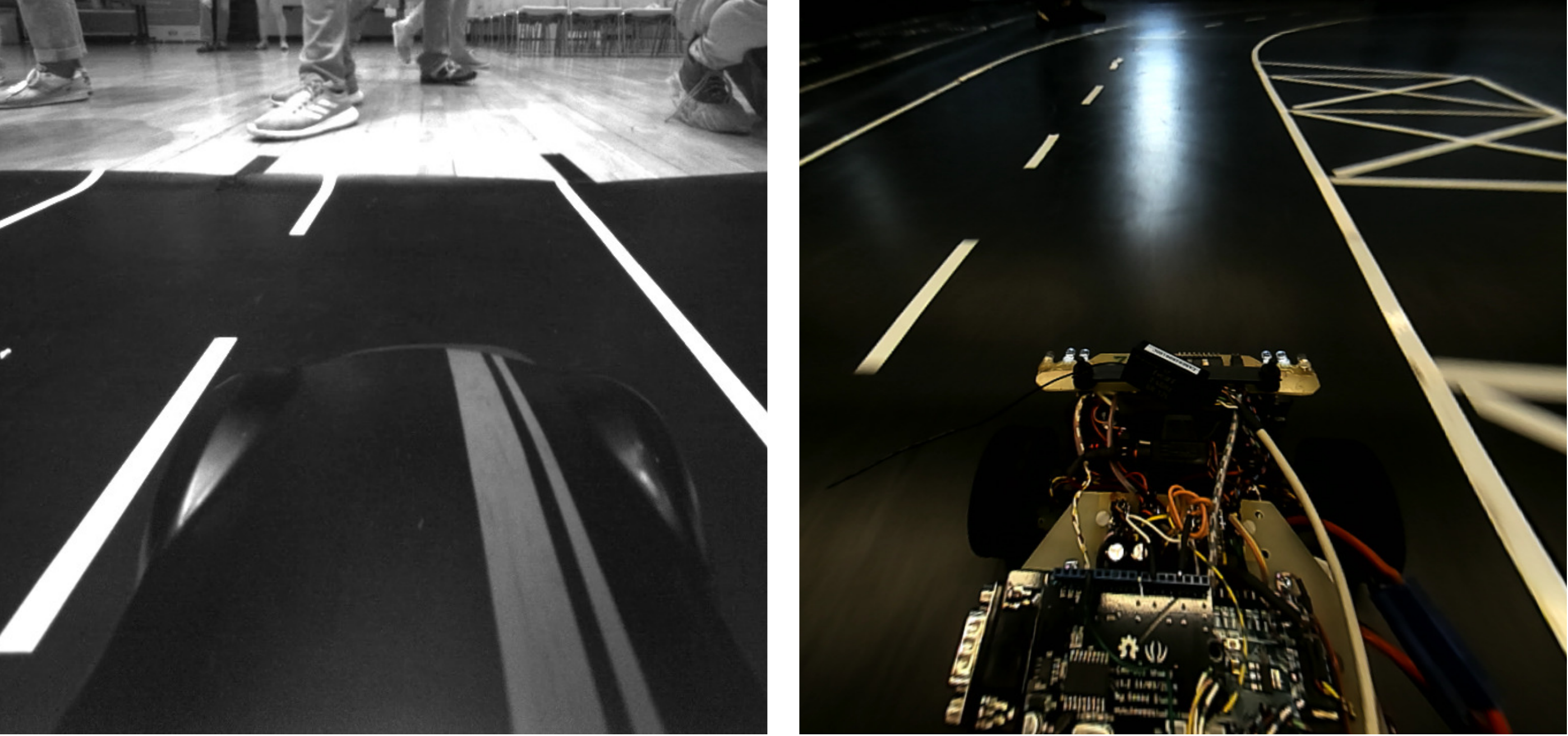}
    \caption{In the left image \cite{k2b3}, surroundings beyond the track like legs, feets and shoes are depicted in the background. The right image \cite{k2b4} shows a reflection on the track caused by natural lightning.}
	\label{fig:real data example}
\end{figure}

To cover all driving scenarios and achieve optimal predictions, synthetic data must be extended with real images. It is necessary to consider special features that are not included in the simulation like natural lightning, blurred scenes, and surroundings beyond the route, as illustrated in Fig.~\ref{fig:real data example}. Hence, real images containing these features are added to the dataset. The real images include different driving maneuvers as well as objects with various orientations and visibility. In addition, depending on the data source, various image resolutions, grayscale and colored images, and various RC vehicles are used \eqref{subsec:Sources of real data}.
%---------------------------------------------------------------------
\subsection{BIRD'S EYE VIEW TRANSFORMATION}
\label{subsec:BirdTrafo}
\begin{figure}[t!]
    \includegraphics[width=.48\textwidth]{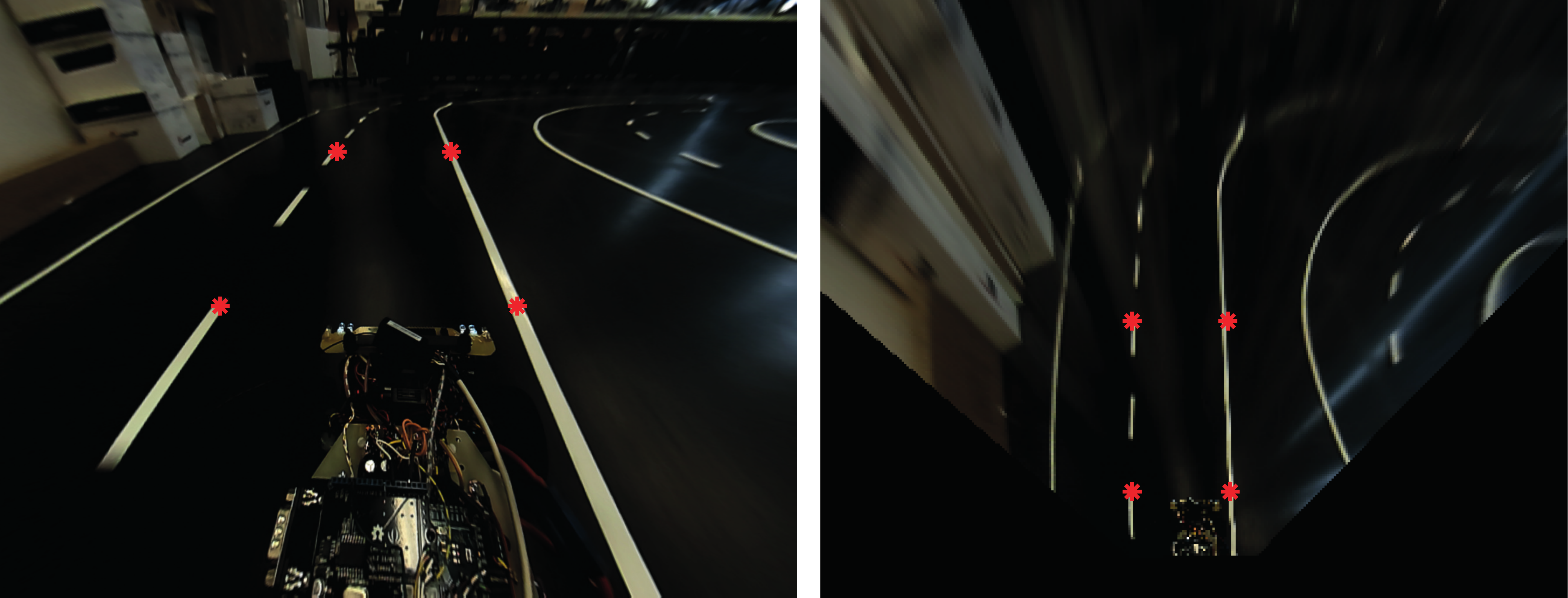}
    \caption{Bird's eye view transformation process using four mapping points illustrated in red.}
	\label{fig:bird_trafo}
\end{figure}

There are multiple ways to transform a first-person perspective image into a bird's eye view perspective. The warp perspective mapping method \cite{y2} is used for this work since no intrinsic nor extrinsic parameters of the cameras are available. This mapping method is suitable for several different camera models and does not require additional calibration. The mapping process consists of selecting four points $X_{ego}$ on the ground plane from the first-person perspective and their corresponding points $X_{bird}$ from the bird's eye view perspective as illustrated in Fig.~\ref{fig:bird_trafo}. $X_{ego}$ of the input image will be viewed as $X_{bird}$. The mapping from $X_{ego}$ to $X_{bird}$ can be expressed as:
\begin{equation}
    X_{bird} = H \cdot X_{ego}
\end{equation}
The transformation matrix $H$ can be calculated using the equation above. The matrix $H$ is then used to map the input images from the first-person to the output of the bird's eye view perspective using a pixel-by-pixel process \cite{y1}.

%+++++++++++++++++++++++++++++++++++++++++++++++++++++++++++++++++++++++++++++++++++++++++++
% Experiments section
\section{Experiments}
In this section, the conducted experiments to test the performance of the proposed model on the generated dataset as well as the respective results are described. In this context, the accuracy of the first-person and bird's eye view perspectives are compared. Also, the frame rate and the accuracy of both perspectives are measured and compared using different model arithmetic on the NVIDIA Jetson AGX Xavier board.
%---------------------------------------------------------------------
\subsection{COMPARISON OF FIRST-PERSON AND BIRD'S EYE VIEW PERSPECTIVE REGARDING ACCURACY}
This experiment compares the accuracy of the semantic segmentation model regarding the first-person and the bird's eye view perspectives. For this purpose, two FasterSeg models are trained using a dataset from the first-person and the bird's eye view perspectives. In the following, the dataset and the hyperparameters, which were used to train the models, are described. Finally, the results of this experiment are presented.\\

\subsubsection{DATASET}
\label{generatedDataset}
\hfill\\
Table \ref{tab:firstdatastat} describes the dataset used for the training of the FasterSeg models. The dataset consists of synthetic and real images, received from the it:movES team. The images are divided into three sets: a training set (Train) containing $75\,\%$ of the images, a validation set (Val) consisting of $25\,\%$ of the images, and a test set (Test) composed of 20 real images used to measure the accuracy of the models. To train the bird's eye view model, all the images are transformed into the bird's eye view perspective using the method described in section \ref{subsec:BirdTrafo}. Both FasterSeg models are trained using the same resolution ($320 \times 256$) to objectively compare both perspectives. It is important to consider that this dataset contains only 11 objects instead of the initial 14 illustrated in Fig.~\ref{fig:annotation definitions}.\\

\begin{table}
    \caption{The dataset generated for the perspective comparison experiment.}
    \setlength{\tabcolsep}{3pt}
    \label{tab:firstdatastat}
        \begin{tabular}{p{50pt}p{50pt}p{52pt}p{20pt}p{20pt}p{15pt}}\toprule    
            \textbf{Data source} &\textbf{Resolution}   &\textbf{Resolution}  &\textbf{Train} &\textbf{Val} &\textbf{Test} \\
                                 &\textbf{first-person perspective} &\textbf{bird's eye view perspective}  &               &             &              \\\midrule
            it:movES (real)                & 320 x 256            & 320 x 256           & 55            & 18          & 20           \\
            it:movES            & 320 x 256            & 320 x 256           & 576           & 192         & 0            \\
            (synthetic)   & & & & & \\\midrule
            \textbf{Sum}         &                      &                     & 631           & 210         & 20           \\\bottomrule
        \end{tabular}
\end{table}
%---------------------------------------------------------------------
\subsubsection{SETTINGS}
\hfill\\
The training process of FasterSeg is divided into four substeps \cite{k29}, \cite{k211}. Note that only the teacher network is used in this experiment. The configured hyperparameters, such as the number of epochs and the number of iterations per epoch, are listed in Table \ref{tab:firsttrainsettings}. The training runs on an NVIDIA Tesla V100S-PCI GPU. %no \\

\begin{table}
    \caption{The hyperparameters adjusted for the perspective comparison experiment.}
    \setlength{\tabcolsep}{3pt}
    \label{tab:firsttrainsettings}
        \begin{tabular}{p{90pt}p{25pt}p{35pt}p{20pt}p{40pt}}\toprule    
            \textbf{Substep} &\textbf{Epochs} &\textbf{Iterations per epoch}&\textbf{Batch size}&\textbf{Initial learning rate}\\\midrule
            Pretrain the supernet & 20 & 400 & 3 & $2\cdot10^{-2}$ \\    
            Search the architecture & 30 & 400 & 2 & $1\cdot10^{-2}$\\
            Train the teacher network & 249 & 1000 & 12 & $1\cdot10^{-2}$\\\bottomrule
        \end{tabular}
\end{table}
%---------------------------------------------------------------------
\subsubsection{RESULTS}
\begin{figure*}[t!]
    \includegraphics[width=\textwidth]{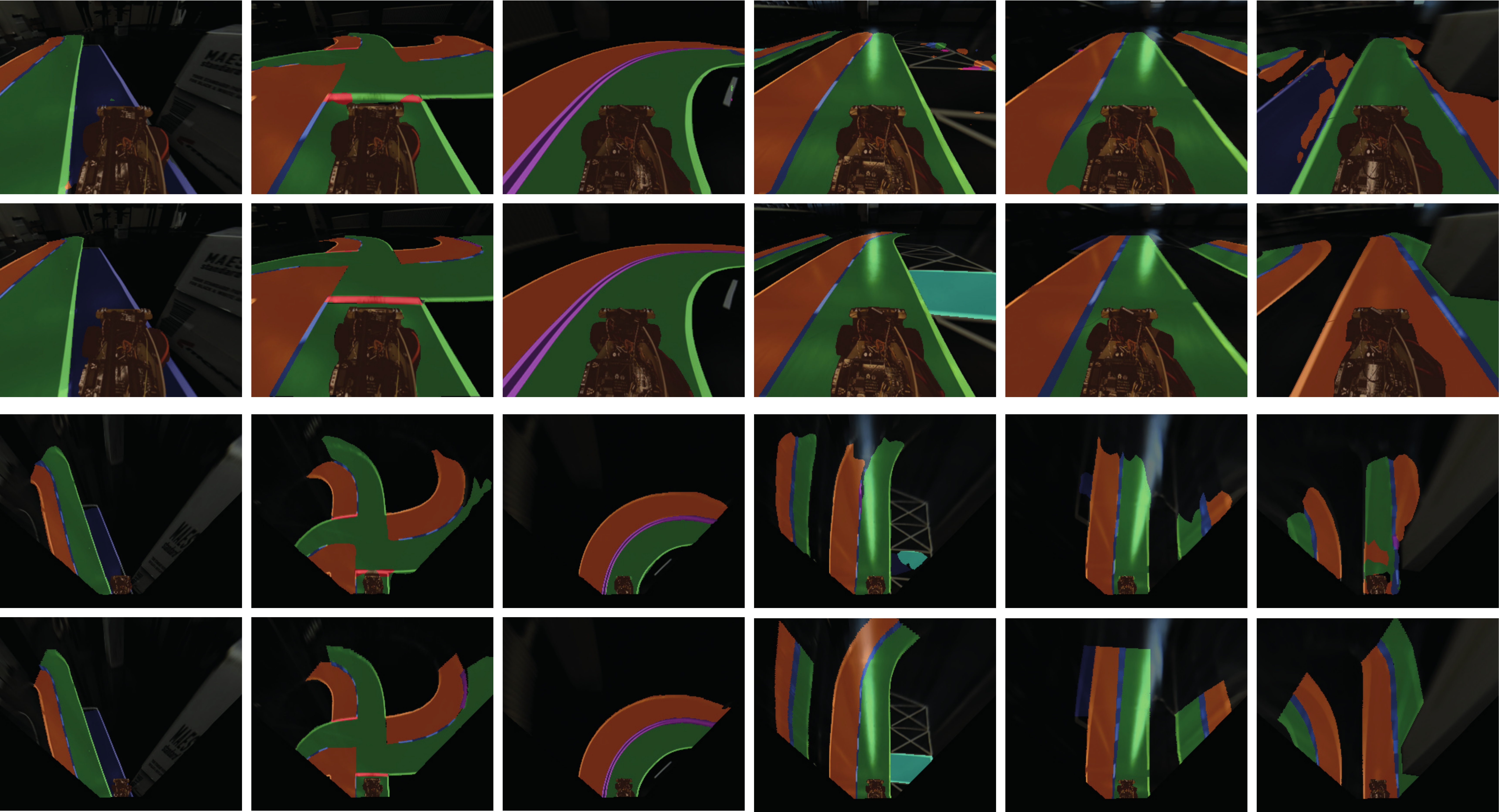}
    \caption{Visual predictions of the trained FasterSeg models using the test dataset. First and second rows show the predictions and the corresponding ground truth images in the first-person perspective.
    Third and fourth rows illustrate the predictions and the corresponding ground truth images in the bird's eye view perspective. Each color corresponds to a predefined class.}
	\label{fig:test predictions it moves}
\end{figure*}

\hfill\\
Table \ref{tab:adjustedegovsbirdview} lists the predictions' accuracies of the trained models. Strikingly, the bird's eye view perspective achieves an mIoU that is almost as good as the first-person perspective. Considering the IoU of each class, the double solid center line and the stop line reach a much higher IoU in the first-person perspective than the bird's eye view perspective. On the other hand, free parking space is predicted better in the bird's eye view perspective. \\ Fig.~\ref{fig:test predictions it moves} shows several test predictions and their corresponding ground truth images. Although some of the images have light reflections on the track, this has no apparent impact on the predictions' quality. Note that the predictions are less accurate when the vehicle changes lanes. The left and the right lanes are often confused during such maneuvers.
Furthermore, both perspectives achieve an inference frame rate of 247.11 FPS on the NVIDIA Jetson AGX Xavier. 

\begin{table}
    \caption{The measured accuracies of the first-person and bird's eye view models tested with uniform resolution.}
    \setlength{\tabcolsep}{3pt}
    \label{tab:adjustedegovsbirdview}
        \begin{tabular}{p{90pt}p{60pt}p{60pt}}\toprule    
            \textbf{Class}           &\textbf{IoU [$\%$]} &\textbf{IoU [$\%$]} \\
                                     &\textbf{first-person}                    &\textbf{bird's eye view}                      \\
                                     &\textbf{perspective}                     &\textbf{perspective}                     \\\midrule
            unlabeled                & 92.74                                   & 96.40                                   \\    
            left lane                & 74.94                                   & 78.32                                   \\
            right lane               & 83.97                                   & 79.43                                   \\   
            dashed center line       & 51.57                                   & 53.84                                   \\
            double solid center line & 89.01                                   & 53.32                                   \\
            starting line            & 65.71                                   & 60.51                                   \\
            stop line                & 50.48                                   & 27.37                                   \\
            crosswalk                & 76.26                                   & 88.89                                   \\
            free parking space       & 7.49                                    & 45.79                                   \\
            free parking area        & 62.27                                   & 56.93                                   \\\midrule
            \textbf{mIoU}            & 65.44                                   & 64.08                                   \\\bottomrule
        \end{tabular}
\end{table}
%---------------------------------------------------------------------
\subsection{REAL-TIME CAPABILITY AND ACCURACY IN RELATION TO THE ARITHMETIC}
This experiment can be divided into two parts. The first part deals with the analysis of the real-time capability of FasterSeg. The second part examines the accuracy of FasterSeg concerning the arithmetic. For this purpose, two models are trained using a dataset different from the experiment above. The dataset consists of various resolution images from the first-person and the bird's eye view perspectives. The inference is conducted on the NVIDIA Jetson AGX Xavier using the FP16 and the FP32 arithmetic. The TensorRT framework is used to perform the inference. In the following, the used dataset and hyperparameters are described. Finally, the results of the experiment are presented.\\

\subsubsection{DATASET}
\hfill\\
The dataset used to train the models is listed in Table \ref{tab:seconddataset}. It consists of synthetic and real images from different data sources with different resolutions. The dataset is also divided into three sets as described in \ref{generatedDataset}. The test set used to compute the accuracy consists of 208 real images. Note that the dataset is significantly larger than the dataset used in the previous experiment. The dataset is also transformed into the bird's eye view perspective to train the second FasterSeg model.\\

\begin{table}
    \caption{The dataset generated for the real-time capability and the arithmetic accuracy comparison experiment.}
    \setlength{\tabcolsep}{3pt}
    \label{tab:seconddataset}
        \begin{tabular}{p{50pt}p{45pt}p{52pt}p{20pt}p{20pt}p{20pt}}\toprule    
            \textbf{Data source} &\textbf{Resolution}   &\textbf{Resolution}  &\textbf{Train} &\textbf{Val} &\textbf{Test} \\
                                 &\textbf{first-person perspective} &\textbf{bird's eye view perspective}  &               &             &              \\\midrule
            it:movES (real)             & 1280 x 960           & 320 x 256           & 48            & 16          & 30           \\
            KITcar               & 1280 x 640           & 320 x 320           & 6             & 2           & 31           \\
            ISF L\"owen          & 768 x 384            & 320 x 320           & 14            & 4           & 69           \\
            Spatzenhirn          & 2048 x 1536          & 256 x 256           & 17            & 6           & 78           \\
            it:movES   & 1280 x 960           & 320 x 256           & 18221         & 6072        & 0            \\
            (synthetic)   & & & & & \\\midrule
            \textbf{Sum}         &                      &                     & 18306         & 6100        & 208          \\\bottomrule
        \end{tabular}
\end{table}

\subsubsection{SETTINGS}
\hfill\\
The adjusted hyperparameters for the FasterSeg models are listed in Table \ref{tab:secondtrainsettings}. Only the teacher network is used for this experiment. The training runs on an NVIDIA Tesla V100S-PCI GPU.\\

\begin{table}
    \caption{The hyperparameters adjusted for the real-time capability and the arithmetic accuracy comparison experiment.}
    \setlength{\tabcolsep}{3pt}
    \label{tab:secondtrainsettings}
        \begin{tabular}{p{90pt}p{25pt}p{35pt}p{20pt}p{40pt}}\toprule    
            \textbf{Substep} &\textbf{Epochs} &\textbf{Iterations per epoch}&\textbf{Batch size}&\textbf{Initial learning rate}\\\midrule
            Pretrain the supernet & 20 & 3051 & 3 & $2\cdot10^{-2}$ \\    
            Search the architecture & 30 & 4576 & 2 & $1\cdot10^{-2}$\\
            Train the teacher network & 420 & 1526 & 12 & $1\cdot10^{-2}$\\\bottomrule
        \end{tabular}
\end{table}

\subsubsection{RESULTS}
\hfill\\
The first part of the experiment consists of measuring the inference frame rate of the bird's eye view and the first-person FasterSeg models using the FP16 and the FP32 arithmetic on an NVIDIA Jetson AGX Xavier board. Table \ref{tab:frame rate Jetson} lists the measurements' results. The results demonstrate the real-time capability of the FasterSeg model on an NVIDIA Jetson AGX Xavier board. The bird's eye view model reaches very high frame rate values, which concludes it is highly dependent on the resolution. Furthermore, the frame rate using FP16 is approximately twice as high as the frame rate with FP32. \\
The second part investigates the impact of the used arithmetic on the achieved accuracy. Therefore, the bird's eye view and the first-person FasterSeg models are compared using the FP16 and the FP32 arithmetic. Table \ref{tab:mIoU Jetson} illustrates the achieved accuracies (mIoU) of the trained models. Note that the dataset used to train the models consists of images with various resolutions from different sources leading to relatively low accuracies. The results demonstrate that the arithmetic has no significant influence on the accuracy.
\begin{table}
    \caption{The frame rate of the bird's eye view and the first-person FasterSeg models using different arithmetic measured on NVIDIA Jetson AGX Xavier.}
    \setlength{\tabcolsep}{3pt}
    \label{tab:frame rate Jetson}
        \begin{tabular}{cp{48pt}p{48pt}p{50pt}p{50pt}}\toprule
            &\textbf{Data source} &\textbf{Resolution} &\textbf{FPS FP16} &\textbf{FPS FP32} \\\midrule
            \multirow{4}{*}{\rotatebox[origin=c]{90}{\textbf{bird's eye}}}
            \multirow{4}{*}{\rotatebox[origin=c]{90}{\textbf{view}}}
            &Spatzenhirn          & 256 x 256          & 319.85           & 190.91           \\
            &it:movES             & 320 x 256          & 274.11           & 153.16           \\
            &KITcar               & 320 x 320          & 231.13           & 118.91           \\
            &ISF L\"owen          & 320 x 320          & 230.81           & 118.68           \\ \midrule
            \multirow{4}{*}{\rotatebox[origin=c]{90}{\textbf{first-}}}
            \multirow{4}{*}{\rotatebox[origin=c]{90}{\textbf{person}}}
            &ISF L\"owen          & 768 x 384          & 113.93           & 55.22            \\
            &KITcar               & 1280 x 640         & 48.66            & 20.87            \\
            &it:movES             & 1280 x 960         & 32.79            & 15.19            \\ 
            &Spatzenhirn          & 2048 x 1536        & 14.27            & 6.39             \\
            \bottomrule
        \end{tabular}
\end{table}

\begin{table}
    \caption{The models accuracies measured on NVIDIA Jetson AGX Xavier.}
    \setlength{\tabcolsep}{3pt}
    \label{tab:mIoU Jetson}
        \begin{tabular}{cp{45pt}p{45pt}p{53pt}p{53pt}}\toprule
            &\textbf{Data source} &\textbf{Resolution} &\textbf{mIoU FP16} &\textbf{mIoU FP32}\\
            &   &   & [$\%$] & [$\%$]\\\midrule
            \multirow{4}{*}{\rotatebox[origin=c]{90}{\textbf{bird's eye}}}
            \multirow{4}{*}{\rotatebox[origin=c]{90}{\textbf{view}}}
            &Spatzenhirn          & 256 x 256          & 28.0206                  & 28.0234                  \\
            &it:movES             & 320 x 256          & 25.7181                  & 25.7040                  \\
            &KITcar               & 320 x 320          & 13.1934                  & 12.1047                  \\
            &ISF L\"owen          & 320 x 320          & 11.8073                  & 11.8127                  \\\midrule
            \multirow{4}{*}{\rotatebox[origin=c]{90}{\textbf{first-}}}
            \multirow{4}{*}{\rotatebox[origin=c]{90}{\textbf{person}}}
            &ISF L\"owen          & 768 x 384          & 44.2960                  & 44.2849                  \\
            &KITcar               & 1280 x 640         & 41.1073                  & 41.1028                  \\
            &it:movES             & 1280 x 960         & 37.3627                  & 37.3606                  \\
            &Spatzenhirn          & 2048 x 1536        & 55.7952                  & 55.7936                  \\\bottomrule
        \end{tabular}
\end{table}

%+++++++++++++++++++++++++++++++++++++++++++++++++++++++++++++++++++++++++++++++++++++++++++
% Conclution section
\section{Conclusion} 
In this paper, the semantic segmentation model FasterSeg was investigated regarding the accuracy and the real-time capability in the Carolo-Cup environment on NVIDIA Jetson AGX Xavier embedded hardware. Synthetic images, which were generated using a semi-automated process, as well as real images were used to train the FasterSeg model. The experimental evaluation demonstrated that FasterSeg model reaches an accuracy of over 64 \% and a frame rate of 247.11 FPS in a Carolo-Cup environment. \\ A process chain was designed to facilitate the generation of labeled synthetic data. This procedure starts with the semi-automated generation of route layouts and trajectories. Images of a modeled Carolo-Cup environment are then recorded, using a Gazebo simulation. Finally, the colored images of the simulation are used to create a suitable annotated dataset for the training. The dataset was also extended by a small amount of manually labeled real images to improve the predictions' accuracy. \\ In addition, the dataset in the first-person perspective was transformed into the bird's eye view perspective to train a separate FasterSeg model. A comparison regarding the accuracy of both perspectives was performed. The comparison showed that both perspectives achieve similar mIoUs. Note that some classes are better predicted in one perspective or the other. Depending on the use case and class relevance, both perspectives provide acceptable predictions. \\ Furthermore, the real-time capability experiment shows that an FP16 arithmetic achieves approximately twice the frame rate compared to an FP32 arithmetic. However, the arithmetic of the inference had no significant impact on the accuracy of the models.

%+++++++++++++++++++++++++++++++++++++++++++++++++++++++++++++++++++++++++++++++++++++++++++
% bibliography


\begin{thebibliography}{00}
    \bibitem{k11} T. Nguyen and M. Yoo, "Fusing LIDAR sensor and RGB camera for object detection in autonomous vehicle with fuzzy logic approach," 2021 International Conference on Information Networking (ICOIN), 2021, pp. 788-791, doi: 10.1109/ICOIN50884.2021.9334015.

    \bibitem{k12} K. Min, S. Han, D. Lee, D. Choi, K. Sung and J. Choi, "SAE Level 3 Autonomous Driving Technology of the ETRI," 2019 International Conference on Information and Communication Technology Convergence (ICTC), 2019, pp. 464-466, doi: 10.1109/ICTC46691.2019.8939765.

    \bibitem{k13} L. Bartolomei, L. Teixeira and M. Chli, "Perception-aware Path Planning for UAVs using Semantic Segmentation," 2020 IEEE/RSJ International Conference on Intelligent Robots and Systems (IROS), 2020, pp. 5808-5815, doi: 10.1109/IROS45743.2020.9341347.

    \bibitem{k14} M. Hua, Y. Nan and S. Lian, "Small Obstacle Avoidance Based on RGB-D Semantic Segmentation," 2019 IEEE/CVF International Conference on Computer Vision Workshop (ICCVW), 2019, pp. 886-894, doi: 10.1109/ICCVW.2019.00117.

    \bibitem{k15} Technische Universit\"at Braunschweig, "Carolo-Master-Cup@Home Regulations 2021," 20 Jan 2021. [Online]. Available: \url{https://www.tu-braunschweig.de/fileadmin/Redaktionsgruppen/Institute\_Fakultaet\_5/Carolo-Cup/Basic-Cup\_Regulations\_210120.pdf}.  [Accessed February 2021].

    \bibitem{k2a1} H. Kirchner, Randomized Road Generation and 3D Visualization, TU Munich. [Online]. Available: \url{https://github.com/tum-phoenix/drive\_sim\_road\_generation}. [Accessed May 2020].

    \bibitem{k2a2} S. Cakir, M. Gau\ss, K. H\"appeler, and Y. Ounajjar, Road generation to use in a simulation for semantic segmentation, Esslingen University of Applied Sciences. [Online]. Available: \url{https://github.com/Sinop97/drive\_sim\_road\_generation}. [Accessed July 2021].

    \bibitem{k2b1} Team Spatzenhirn, University of Ulm. [Online]. Available: \url{https://www.uni-ulm.de/en/in/spatzenhirn/}. [Accessed July 2021].

    \bibitem{k2b2} Team ISF L\"owen, Institute of Software Engineering and Automotive Informatics, TU Braunschweig. [Online]. Available: \url{http://www.isf-loewen.de/}. [Accessed July 2021].

    \bibitem{k2b3} Team KITcar, Karlsruhe Institute of Technology. [Online]. Available: \url{https://kitcar-team.de/}. [Accessed July 2021].

    \bibitem{k2b4} Team it:movES, Esslingen University of Applied Sciences. [Online]. Available: \url{https://www.hs-esslingen.de/informatik-und-informationstechnik/forschung-labore/projekte/interne-projekte/}. [Accessed July 2021].

    \bibitem{k2c1} Lin, T. Y., Maire, M., Belongie, S., Hays, J., Perona, P., Ramanan, D., ... \& Zitnick, C. L. (2014, September). Microsoft coco: Common objects in context. In European conference on computer vision (pp. 740-755). Springer, Cham

    \bibitem{k2c2} M. Cordts, M. Omran, S. Ramos, T. Rehfeld, M. Enzweiler, R. Benenson, U. Franke, S. Roth, and B. Schiele, ``The Cityscapes Dataset for Semantic Urban Scene Understanding,'' 2016 IEEE Conference on Computer Vision and Pattern Recognition (CVPR), 2016, pp. 3213-3223, doi: 10.1109/CVPR.2016.350.

    \bibitem{k21} H. Chen, K. Sun, Z. Tian, C. Shen, Y. Huang and Y. Yan, "BlendMask: Top-Down Meets Bottom-Up for Instance Segmentation," 2020 IEEE/CVF Conference on Computer Vision and Pattern Recognition (CVPR), 2020, pp. 8570-8578, doi: 10.1109/CVPR42600.2020.00860.

    \bibitem{k22} D. Bolya, C. Zhou, F. Xiao and Y. J. Lee, "YOLACT++ Better Real-Time Instance Segmentation," in IEEE Transactions on Pattern Analysis and Machine Intelligence, vol. 44, no. 2, pp. 1108-1121, 1 Feb. 2022, doi: 10.1109/TPAMI.2020.3014297

    \bibitem{k23} Y. Lee and J. Park, "CenterMask: Real-Time Anchor-Free Instance Segmentation," 2020 IEEE/CVF Conference on Computer Vision and Pattern Recognition (CVPR), 2020, pp. 13903-13912, doi: 10.1109/CVPR42600.2020.01392.

    \bibitem{k24} D. Bolya, C. Zhou, F. Xiao and Y. J. Lee, "YOLACT: Real-Time Instance Segmentation," 2019 IEEE/CVF International Conference on Computer Vision (ICCV), 2019, pp. 9156-9165, doi: 10.1109/ICCV.2019.00925.

    \bibitem{k25} P. Chao, C. -Y. Kao, Y. Ruan, C. -H. Huang and Y. -L. Lin, "HarDNet: A Low Memory Traffic Network," 2019 IEEE/CVF International Conference on Computer Vision (ICCV), 2019, pp. 3551-3560, doi: 10.1109/ICCV.2019.00365.

    \bibitem{k26} Papers with code, ``HarDNet: A Low Memory Traffic Network,'' Accessed: Jul 01, 2021. [Online]. Available: \url{https://paperswithcode.com/paper/hardnet-a-low-memory-traffic-network}

    \bibitem{k27} M. Or\v{s}ic, I. Kre\v{s}o, P. Bevandi\'{c}, and S. \v{S}egvi\'{c}, ``In Defense of Pre-trained ImageNet Architectures for Real-time Semantic Segmentation of Road-driving Images,'' Apr. 12, 2019, arXiv:1903.08469v2. [Online]. Available: \url{https://arxiv.org/pdf/1903.08469v2.pdf}

    \bibitem{k28} C. Yu, C. Gao, J. Wang, G. Yu, C. Shen and N.Sang, ``BiSeNet V2: Bilateral Network with Guided Aggregation for Real-time Semantic Segmentation,'' Apr. 5, 2020, arXiv:2004.02147v1. [Online]. Available: \url{https://arxiv.org/pdf/2004.02147v1.pdf}

    \bibitem{k29} W.Chen, X. Gong, X. Liu, Q. Zhang, Y. Li and Z. Wang, ``FASTERSEG: SEARCHING FOR FASTER REAL-TIME SEMANTIC SEGMENTATION,'' Jan. 16, 2020, arXiv:1912.10917. [Online]. Available: \url{https://arxiv.org/pdf/1912.10917.pdf}

    \bibitem{k210} C. Yu, J. Wang, C. Peng, C. Gao, G. Yu, and N.Sang. (2018). Bisenet: Bilateral segmentation network for real-time semantic segmentation. In Proceedings of the European conference on computer vision (ECCV) (pp. 325-341).

    \bibitem{k211} S. Cakir, M. Gau\ss, K. H\"appeler, and Y. Ounajjar, How to train FasterSeg for custom objects, Esslingen University of Applied Sciences. [Online]. Available: \url{https://github.com/Gaussianer/FasterSeg}. [Accessed July 2021].

    \bibitem{y1} DUONG, Tin Trung, PHAM, Cuong Cao, TRAN, Tai Huu-Phuong, et al. Near real-time ego-lane detection in highway and urban streets. In : 2016 IEEE International Conference on Consumer Electronics-Asia (ICCE-Asia). IEEE, 2016. p. 1-4.

    \bibitem{y2} KIM, ZuWhan. Robust lane detection and tracking in challenging scenarios. IEEE Transactions on intelligent transportation systems, 2008, vol. 9, no 1, p. 16-26.

    \bibitem{k30} S. Minaee, Y. Y. Boykov, F. Porikli, A. J. Plaza, N. Kehtarnavaz and D. Terzopoulos, "Image Segmentation Using Deep Learning: A Survey," in IEEE Transactions on Pattern Analysis and Machine Intelligence, doi: 10.1109/TPAMI.2021.3059968

    \bibitem{k31} Everingham, M., Eslami, S.M.A., Van Gool, L. et al. The Pascal Visual Object Classes Challenge: A Retrospective. Int J Comput Vis 111, 98?136 (2015). \url{https://doi.org/10.1007/s11263-014-0733-5}
\end{thebibliography}
\end{document}